\documentclass[runningheads]{llncs}
\usepackage[T1]{fontenc}
\usepackage{graphicx}
\usepackage{booktabs}
\usepackage[misc]{ifsym}
\newcommand{\corr}{(\Letter)}
\usepackage{mwe}

\usepackage{cite}
\usepackage{amsmath,amssymb,amsfonts}
\usepackage{algorithmic}
\usepackage{graphicx}
\usepackage{textcomp}
\usepackage{xcolor}
\usepackage{booktabs}
\usepackage{subcaption}

\begin{document}

\title{SODA: Out-of-Distribution Detection in Domain-Shifted Point Clouds via Neighborhood Propagation}

\titlerunning{SODA: OOD Detection in Domain-Shifted Point Clouds}

\author{Adam Goodge\inst{1} \and
Xun Xu\inst{1} \corr \and
Bryan Hooi\inst{2} \and
Wee Siong Ng\inst{1}\and
Jingyi Liao\inst{1,2} \and
Yongyi Su\inst{1}\and
Xulei Yang\inst{1}
}

\authorrunning{A. Goodge et al.}

\institute{Institute for Infocomm Research, Agency for Science, Technology and Research (A*STAR), Singapore \\ \email{xu\_xun@i2r.a-star.edu.sg} \and
School of Computing, National University of Singapore, Singapore}

\maketitle              

\begin{abstract}
As point cloud data increases in prevalence in a variety of applications, the ability to detect out-of-distribution (OOD) point cloud objects becomes critical for ensuring model safety and reliability. However, this problem remains under-explored in existing research. Inspired by success in the image domain, we propose to exploit advances in 3D vision-language models (3D VLMs) for OOD detection in point cloud objects. However, a major challenge is that point cloud datasets used to pre-train 3D VLMs are drastically smaller in size and object diversity than their image-based counterparts. Critically, they often contain exclusively computer-designed synthetic objects. This leads to a substantial domain shift when the model is transferred to practical tasks involving real objects scanned from the physical environment. In this paper, our empirical experiments show that synthetic-to-real domain shift significantly degrades the alignment of point cloud with their associated text embeddings in the 3D VLM latent space, hindering downstream performance. To address this, we propose a novel methodology called SODA which improves the detection of OOD point clouds through a neighborhood-based score propagation scheme. SODA is inference-based, requires no additional model training, and achieves state-of-the-art performance over existing approaches across datasets and problem settings.

\end{abstract}

\section{Introduction}\label{sec:intro}
Out-of-distribution (OOD) detection is crucial for ensuring the safe and reliable deployment of models in practical applications. It has received significant research attention for 2D images, but remains largely unexplored in the context of 3D point clouds. The detection of unknown 3D objects poses a critical challenge in important applications including robotics, autonomous driving, and healthcare.

Vision-language models (VLMs), such as CLIP \cite{radford2021learning}, have proven effective at OOD detection in image data. OOD inputs are detected by their low cosine similarity to the text embeddings of in-distribution (ID) class labels. Recently, several 3D VLMs incorporating point clouds into this multi-modal latent space have emerged. These models demonstrate strong performance in point cloud classification, but have yet to be evaluated for OOD detection. 

\begin{figure}[h!]
    \centering
    \includegraphics[width=0.75\linewidth]{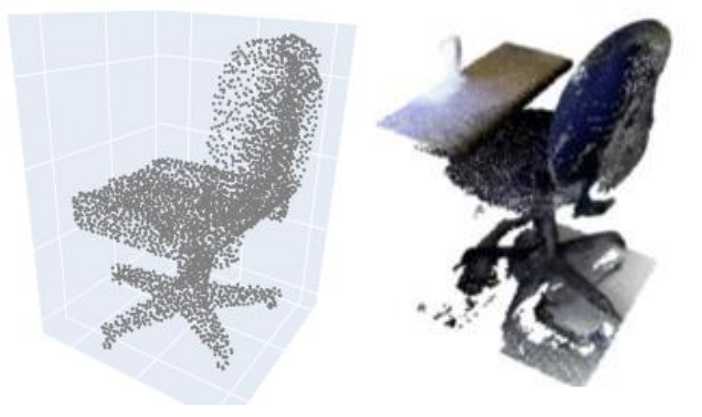}
    \caption{Example of a synthetic chair sample (left) and real chair sample (right).}
    \label{fig:teaser_image}
\end{figure}

A major advantage of VLMs is that they only require ID class labels, not labeled examples of ID data, due to their extensive pre-training on broad data. This is particularly useful in the context of point cloud data, which are typically more expensive to collect and annotate than images. However, point cloud datasets used to pre-train 3D VLMs are also much more limited in size and object diversity than their image counterparts. Moreover, they typically contain only synthetic, computer-designed objects. This results in domain shift when the model is transferred to practical downstream tasks that involve real objects scanned from the physical environment. Figure \ref{fig:teaser_image} shows a synthetic chair from ShapeNet \cite{chang2015shapenet} and a real chair from ScanObjectNN \cite{uy-scanobjectnn-iccv19}. Real data can diverge from synthetic data due to several phenomena, such as: non-uniform point density, sensor noise, occlusion, background objects and physical imperfections. As most practical tasks involve real data, it is crucial that models are robust to this domain shift if they are to be deployed in practice.

In this paper, we show through empirical experiments that ``\textit{synthetic-to-real}" domain shift hinders the alignment of the embeddings of real point clouds with their text labels in the latent space of 3D VLMs, evidenced by the degradation in classification performance of real objects. Moreover, we observe that the severity of this degradation has a strong correlation with the severity of the domain shift. However, we also observe that the model is effective at clustering real data by class in the latent space, suggesting that the model maintains an ability to distinguish between different classes of real objects. These observations motivate our proposed methodology for point cloud OOD detection, called \textbf{S}coring for \textbf{O}ut-of-Distribution \textbf{D}etection through \textbf{A}ggregation (\textbf{SODA}). We use the similarity of point clouds to ID class labels to initialize OOD scores, followed by an important score refinement step based on a neighborhood-based score propagation scheme which accounts for the severity of domain shift through dynamic `source-similarity' weighting. We conduct comprehensive experimentation to evaluate our approach and find that it significantly improves OOD detection without additional fine-tuning of the backbone model.

In summary, our contributions are as follows:
\begin{itemize}
    \item We investigate the effect of synthetic-to-real domain shift on 3D VLMs and find a degradation in the alignment of point cloud and text embeddings.
    \item We propose SODA, a novel methodology to improve OOD detection in domain shifted point-clouds via domain shift-aware neighborhood propagation.
    \item We show that this approach achieves state-of-the-art performance through rigorous experiments and ablation study.
\end{itemize}
The code to reproduce our experiments is available online. 

\section{Related Work}

Numerous methods have been proposed for OOD detection in image data. Notably, MSP \cite{hendrycks2016baseline} uses the maximum softmax probability assigned to a sample by a classification model trained on ID classes, and subsequent methods adopt a similar approach using different confidence measures \cite{liang2017enhancing,liu2020energy,huang2021importance,sun2021react}. However, neural networks are known to make overly confident predictions even for OOD data \cite{guo2017calibration}. OOD point cloud detection remains comparatively  under-explored, and existing studies mostly focus on adapting established methods to point-cloud feature extraction models such as PointNet++ \cite{qi2017pointnet++} and DGCNN \cite{wang2019dynamic} in \cite{alliegro20223dos}, and PointPillars \cite{lang2019pointpillars} in \cite{huang2022out}, VAEs \cite{masuda2021toward}, and teacher-student models \cite{bhardwaj2021empowering}. OpenPatch \cite{rabino2023openpatch} measures the distance of test sample patches to a memory bank containing patches of ID samples.

Recently, strong performance has been achieved in image-based OOD detection with vision-language models \cite{esmaeilpour2022zero,fort2021exploring,wang2023clipn,goodge2024text}. VLMs are typically trained on massive-scale image datasets that encompass a diverse range of classes, domains, and environments. This extensive exposure enables the model to learn rich visual representations, enhancing its ability to generalize across domains. In the 3D VLM space, two approaches have emerged. One approach is to project point cloud data into image inputs for existing 2D VLMs, examples of which include PointCLIP \cite{zhang2022pointclip} and CLIP2Point \cite{huang2023clip2point}. Another approach is to introduce an additional point cloud encoder and train it to align point cloud embeddings with their paired image and text inputs extracted from a fixed 2D VLM. ULIP \cite{xue2023ulip} and ULIP-2 \cite{xue2023ulip2} are notable examples, and the latter achieves state-of-the-art performance in downstream tasks.

As mentioned, point cloud datasets used to pre-train 3D VLMs are relatively limited in size and object diversity, limiting their generalization. Significant research attention has focused on domain adaptation to real point cloud data \cite{huch2023towards}. Common approaches include domain-invariant training or fine-tuning with task-specific data. In an OOD setting, this means adapting to ID data for each task individually, which is cumbersome or even infeasible in data-scarce scenarios. It also risks worsening performance through catastrophic forgetting \cite{rabino2023openpatch}. As such, we focus on leveraging the pre-trained embedding space for training-free adaptation to synthetic-to-real domain shift, without adjusting model parameters.


\section{Motivation}\label{sec:motivation}



In this section, we investigate the effect of synthetic-to-real domain shift on the embeddings learnt by 3D VLMs. In our analysis, we use synthetic samples from ModelNet40\cite{wu20153d} and real samples from ScanObjectNN \cite{uy-scanobjectnn-iccv19}. We focus on only the classes which are common to both datasets (listed under SR1 and SR2 in Table \ref{tab:class_splits}) in order to focus on the effect of domain shift and not the semantic differences between classes. We use ULIP-2 \cite{xue2023ulip2} as our fixed backbone VLM due to its superior performance. ULIP-2 trains a PointBERT point cloud encoder to align point cloud embeddings with paired image and text caption embeddings learnt by a 2D VLM. We extract the embeddings of both synthetic and real point clouds and make several observations:

\begin{table}[h!]
    \centering
    \setlength{\tabcolsep}{24pt}
    \renewcommand{\arraystretch}{1} 
    \caption{Class splits for experiments with ScanObjectNN. Classes under SR1 and SR2 overlap with classes of the same name in ModelNet40, but classes under SR3 classes do not.}
    \begin{tabular}{ccc}
    \toprule
    SR1 & SR2 & SR3 \\ 
    \midrule
    chair & bed & bag \\ 
    shelf & toilet & bin \\ 
    door & desk & box \\ 
    sink & table & pillow \\ 
    sofa & display & cabinet \\ 
    \bottomrule
    \end{tabular}
    \label{tab:class_splits}
\end{table}

\subsubsection{Observation 1: Degraded text-point cloud alignment}
\begin{figure}
    \centering
    \includegraphics[width=0.9\linewidth]{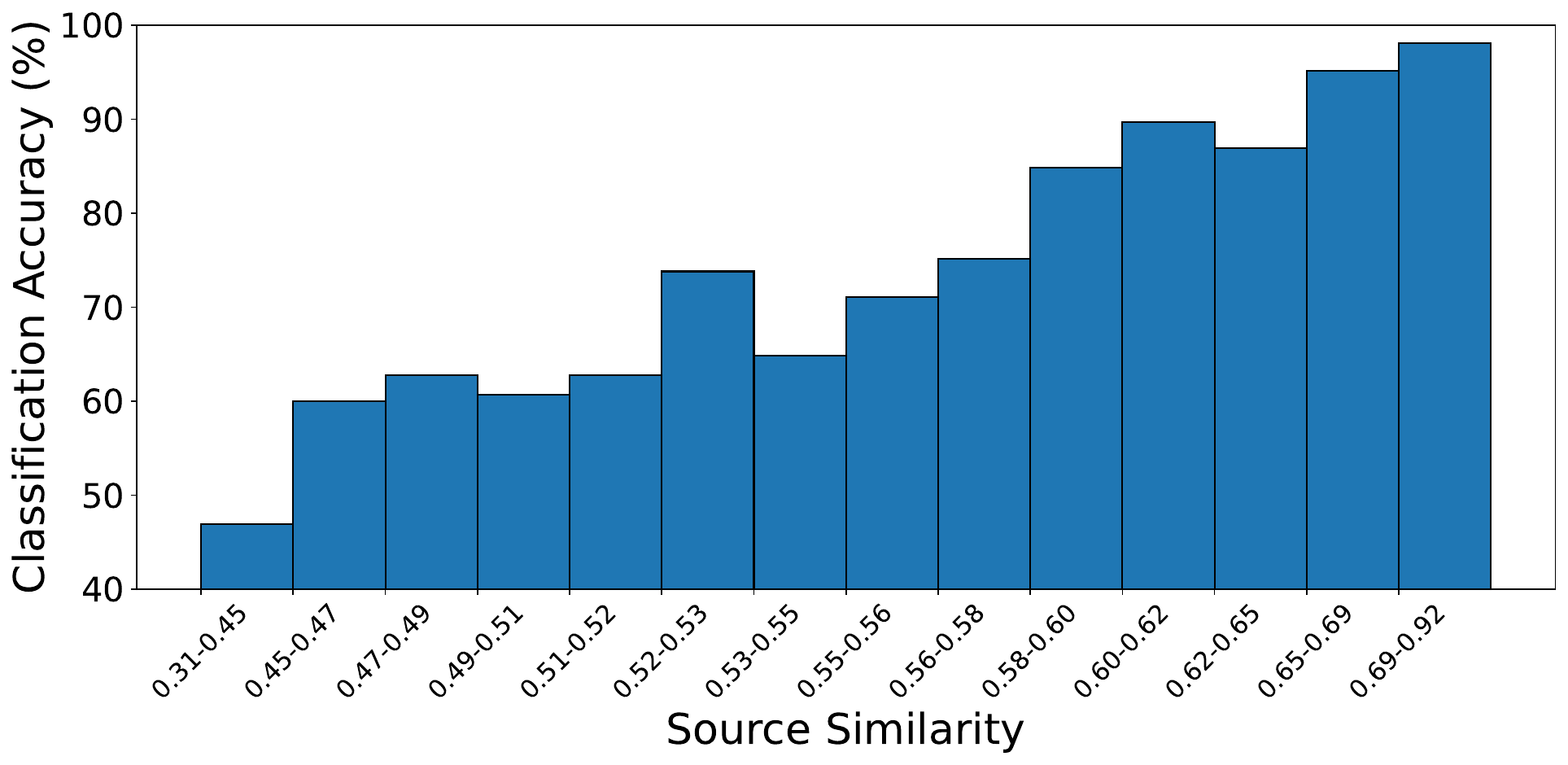}
    \caption{Classification accuracy (\%) of real ScanObjectNN point clouds sorted by cosine similarity to synthetic ModelNet40 samples.}
    \label{fig:bins}
\end{figure}

Following \cite{xue2023ulip2}, we obtain text embeddings for each class label by creating class-based text prompts from a set of templates (more details in Section \ref{sec:SODA}). We classify each sample according to their maximum cosine similarity to these text embeddings. With this setup, we observe that \textbf{classification accuracy drops from 94.27\% on synthetic data to 73.83\% on real data}. This suggests a significant degradation in the alignment of real data with their associated text labels in the latent space compared with synthetic data. This is detrimental to OOD detection, as OOD data may be inappropriately matched to ID class labels, resulting in false negative errors, and vice versa. However, this degradation is not uniform. We take the average similarity of real samples to their 10 nearest synthetic samples as a measure of how `close' they are to the source domain. In Figure  \ref{fig:bins}, we see that this `source similarity' has a direct, linear relationship with classification accuracy. In other words, \textbf{real samples that are closer to the source domain are better aligned with the text labels}.

\subsubsection{Observation 2: Strong class-based clustering}
\begin{figure}[h!]
    \centering
    \includegraphics[width=0.6\linewidth, height = 6.5cm]{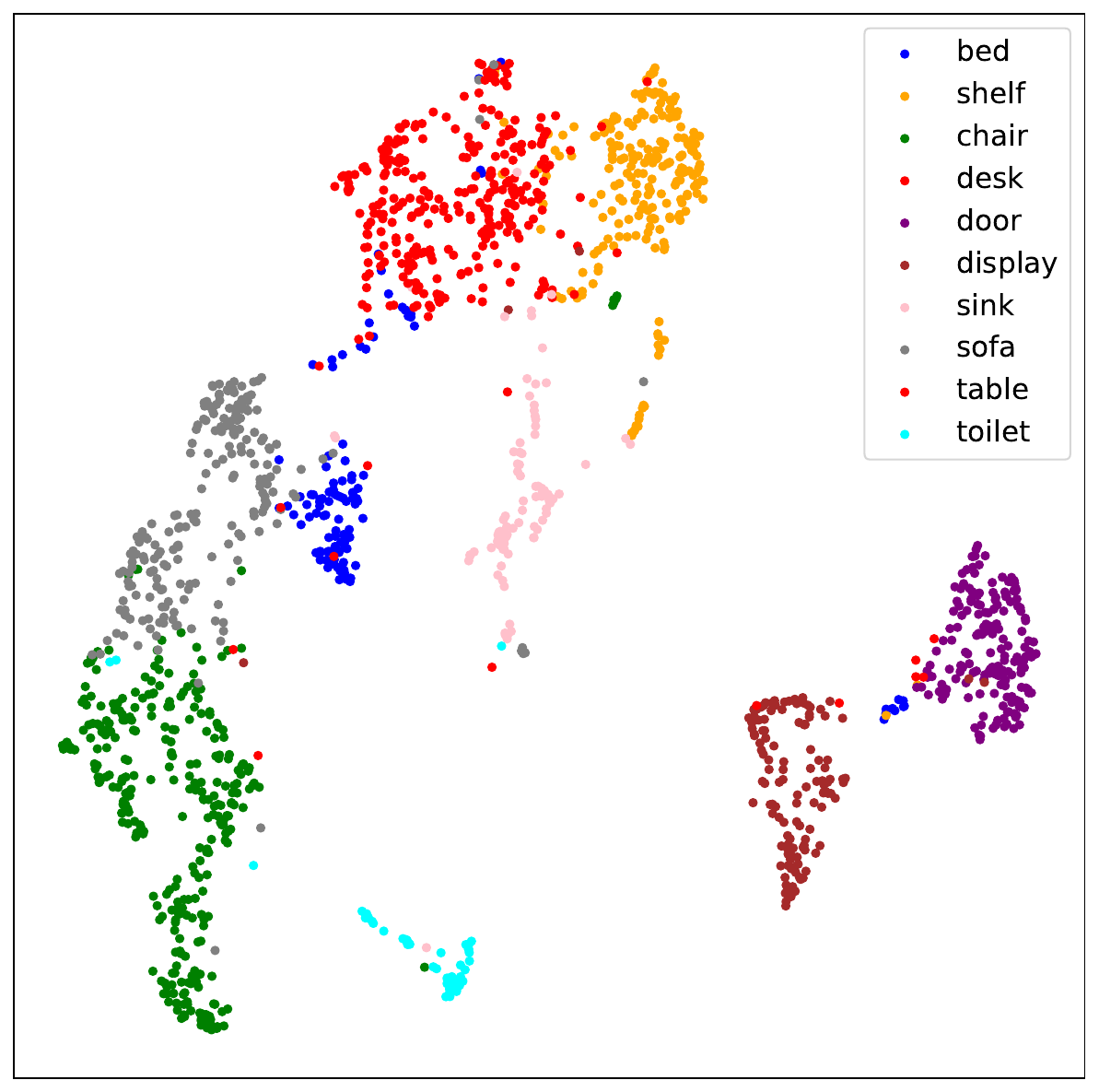}
    \caption{UMAP projections of real domain test point clouds colored by their class.}
    \label{fig:real-tsne}
\end{figure}

Our second observation is that real samples are strongly clustered by class. Figure \ref{fig:real-tsne} visualises the UMAP projections \cite{mcinnes2018umap} of real point samples, with their colour corresponding to their class label. 
We see that the vast majority of samples are well clustered according to class. This suggests that, despite a weakened alignment with the text embeddings, the model retains an ability to distinguish between examples of different classes of real objects. This is promising for OOD detection, as it means that we can expect a real sample from an ID class to be mostly neighbored by other samples of the same ID class, and vice versa for OOD samples. These observations provide motivation for our proposed methodology. 

\section{Methodology}

\begin{figure*}
    \centering
    \includegraphics[width=\linewidth]{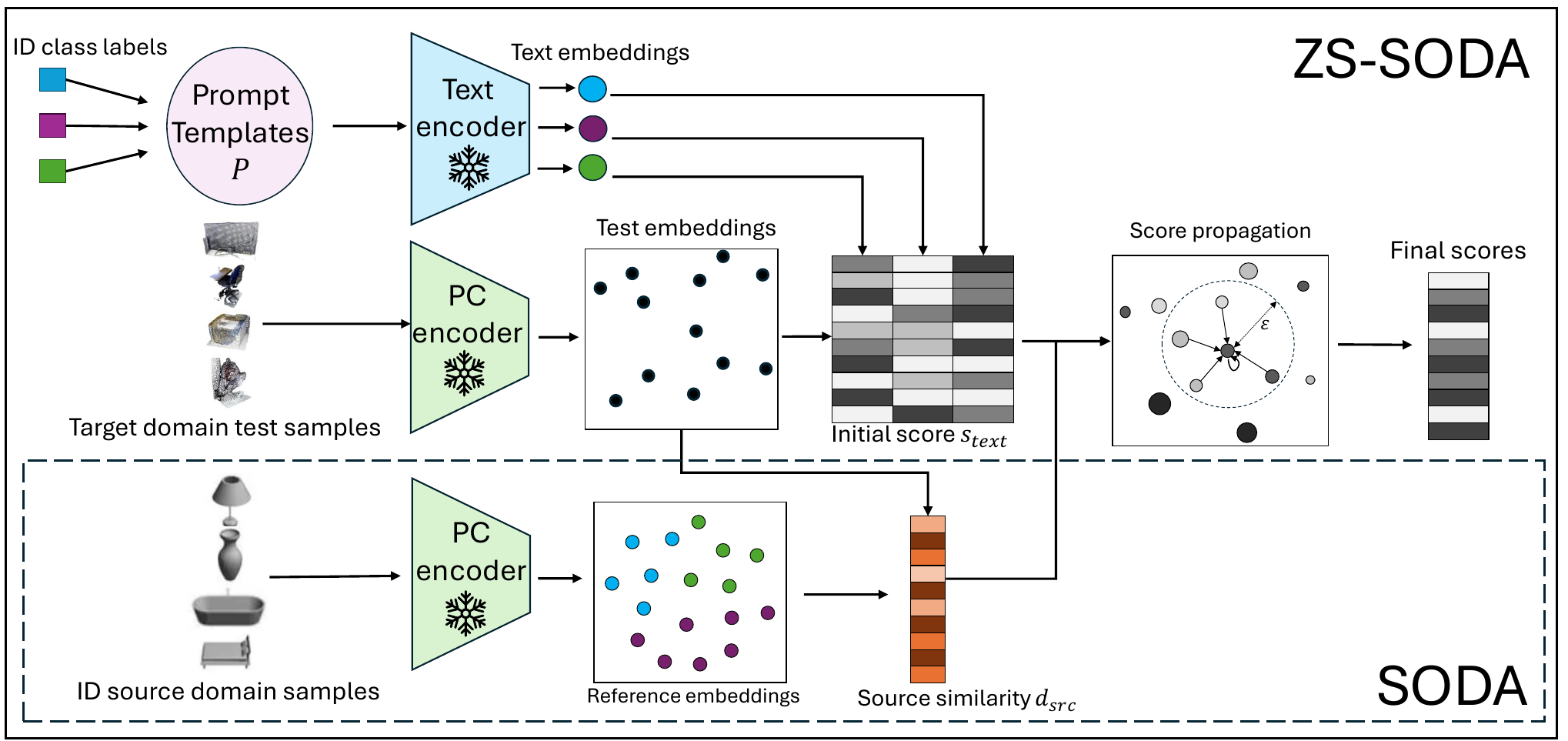}
    \caption{An overview of SODA. Test samples are assigned initial scores based on their similarity to the ID class text embeddings. If available, these scores are multiplied by their source similarity, and refined via neighborhood propagation within an $\varepsilon$-similarity radius. Target domain images are ScanObjectNN and source domain images are ModelNet40.}
    \label{fig:pipeline}
\end{figure*}

Our work most closely relates to \cite{yang2021exploiting}, which exploits neighborhood structure within the test set for classification. However, a major difference in OOD detection is that we do not know the label space of OOD samples, which can belong to any unknown class, therefore it is a more challenging open-set problem. Our methodology is inspired by the label propagation algorithm \cite{zhu2003semi}, a semi-supervised approach to classification which propagates class information from labeled to unlabeled data. However, as all test data is unlabeled in our setting, we have no ground truth information to propagate. \cite{wu2023energy} boosts the detection of OOD nodes within a graph by propagating energy-based scores between neighboring nodes. We explore the potential of this framework beyond graph data, particularly for addressing the challenges of domain shift in non-structured data, exploiting the strong class-based latent clustering of 3D VLMs to improve OOD detection performance in a transductive setting. Our methodology is illustrated in Figure \ref{fig:pipeline}.

\subsection{Problem Statement} We have a set of $N$ ID class labels $\mathcal{C} = \{C_1, C_2, ..., C_N\}$, and an unlabeled test set $\mathbf{X}$, consisting of both ID and OOD real data. A sample is ID if it belongs to any ID class $C_i \in \mathcal{C}$ and OOD otherwise. We aim to distinguish between ID and OOD samples by devising a scoring function which assigns higher scores to ID samples than OOD samples in $\mathbf{X}$.

\subsection{SODA}\label{sec:SODA}

We begin by describing the source-free, zero-shot version of SODA (ZS-SODA), and then explain how source domain samples are incorporated to enhance performance in our full SODA methodology. 

\subsubsection{Initial Scoring}

We use a fixed, pre-trained 3D VLM with text encoder $g_t$ and point cloud encoder $g_{pc}$. Following convention, we use a set of templates $\mathcal{P}$, which are shared between all classes, combined with each class label $C_i$ to create a set of class-specific text prompts $\mathcal{P}_i$. An example template is ``\textit{a point cloud model of a CLS}" where \textit{CLS} is replaced with $C_i$, and the other templates can be found in the supplementary material. These text prompts are encoded and $l2$-normalized, and we take the mean of these embeddings as the text prototype of class $C_i$:
\begin{align}
    \mathbf{P}_i = \frac{1}{|\mathcal{P}_{i}|} \sum_{p \in \mathcal{P}_{i}} g_{t}(p), \  \ i\in1,..,N.
\end{align}
Similarly, we obtain the embedding of a query test sample $\mathbf{x}_i$ via the fixed point cloud encoder: $\mathbf{z}_i =  g_{pc}(\mathbf{x}_i)$. With this, we assign the initial score of $\mathbf{x}_i$ as the maximum of its cosine similarities to the text prototypes:
    \begin{align}
    s_{text}(\mathbf{x}_i) = \max_{i = 1,...,N} \left(sim(\mathbf{z}_i,\mathbf{P}_i)\right)
    \end{align}
ID samples are assumed to have higher similarity to their correct class label, resulting in a high score, compared with OOD samples. However, as we observed, the alignment of point cloud embeddings with their text embeddings is degraded by synthetic-to-real domain shift, which challenges this assumption. To address this, we introduce a score refinement step through neighborhood-based score propagation.

\subsubsection{Score Propagation}
The misalignment of real point cloud embeddings with the text embeddings means that the initial scores are prone to substantial uncertainty or noise. However, we also observed that nearby samples are likely to belong to the same class. Based on this, we refine initial scores using the scores of nearby samples. For example, an ID sample may have an erroneously low score, but by adjusting its score in line with its neighbors, this uncertainty is reduced, the score landscape is smoothened over the latent space, and the overall robustness of OOD scores of individual samples is improved. We achieve this by constructing a \textit{similarity graph}, with a node for each test sample and an edge connecting nodes $i$ and $j$ if their cosine similarity is greater than $\varepsilon$:
\begin{align}\label{eq:threshold}
e_{i,j} = \begin{cases}
    1, & \text{if} \ sim(\mathbf{z_i},\mathbf{z_j}) \geq \varepsilon\\
    0, & \text{otherwise}
\end{cases}
\end{align}
We set $\varepsilon = \text{percentile}(\mathbf{S}, 100(1-\eta))$, where $\mathbf{S}$ is the similarities between all pair-wise test samples and $\eta$ is a hyper-parameter. A smaller $\eta$ results in a higher $\varepsilon$ threshold and a sparser similarity graph, i.e. fewer neighbors per sample on average. This formulation is adaptive to the local density of each sample; densely packed samples will have more neighbors within an $\varepsilon$-similarity radius and therefore more neighbors in the similarity graph, compared to samples in sparse regions. This adaptive strategy is important to limit the propagation of information to only the most similar neighbors, rather than an inflexible $k$ nearest neighbor strategy. We set $\eta = 0.02$ in our main experiments and our ablation study shows that performance is robust within a reasonable range of $\eta$. 

We update the score of a sample $\mathbf{x}_i$ as follows:
\begin{align}\label{eq:propagation}
s^{(t)}(\mathbf{x}_i) = \alpha s^{(0)}(\mathbf{x}_i) + \frac{1-\alpha}{|\mathcal{N}_i|}\sum_{j \in \mathcal{N}_{i}} s^{(t-1)}(\mathbf{x}_j),
 \end{align}
where $t$ is the current iteration, $s^{(0)}(\mathbf{x}_i) = s_{text}(\mathbf{x}_i)$ and $\mathcal{N}_{i}$ is the set of neighbors to node $i$ in the similarity graph. In other words, its updated score is a weighted sum of its initial score and the mean of its neighbors' current scores. Through this procedure, the score of an individual sample is moved closer to those of its neighbors, while anchored to its own initial score to avoid instability, which smoothens the scoring function over the local neighborhood. We allow self-loops in the graph so that every node has at least one neighbor ($|\mathcal{N}_{i}|>0\  \forall i$). We complete $T$ iterations to obtain the final OOD scores. $T$ is a hyper-parameter, and we experiment with different $T$ settings in the ablation study.


\subsection{Source Similarity}

We observed that real samples that are closer to the source domain in cosine similarity are more likely to be correctly classified through text-based similarity matching. Based on this observation, we hypothesise that these samples are better aligned with the text labels, which means that their text-based OOD scores are more reliable. As such, in our full SODA methodology, we use this source similarity to re-weight the importance of different neighbors during score propagation. More formally, given a set of `reference samples' from ID classes in the source domain, $\mathbf{X}^{ref}$, we extract the embeddings of each reference sample $\mathbf{x}^{ref}_i$ from the same fixed point cloud encoder: $\mathbf{z}^{ref}_i = g_{pc}(\mathbf{x}^{ref}_i).$ For a test sample $\mathbf{x}_i$, we take the mean of the cosine similarities of $\mathbf{z}_i$ to its top$k$ ($k = 10$) closest reference sample embeddings to define its source similarity, denoted $d_{src}$:
\begin{align}\label{eq:reference score}
    d_{src}(\mathbf{x}_i) =  \frac{1}{k}\sum_{j \in \text{top}k(\mathbf{z}_i)} (sim(\mathbf{z}_j^{ref}, \mathbf{z}_i))
\end{align}
To further benefit from neighborhood propagation, we iteratively update $d_{src}$ for each sample according to the same formula as Eq. \ref{eq:propagation}, with  $s^{(0)} = d_{src}$. With this, the updated OOD score at the $t^{th}$ iteration is given by:
\begin{equation}\label{eq:combined_score}
    s^{(t)}(\mathbf{x}_i) = d_{src}^{(t)}(\mathbf{x}_i) \cdot s_{text}^{(t)}(\mathbf{x}_i).
\end{equation}
The effect of this is two-fold. Firstly, the OOD scores of samples with high $d_{src}$ will increase. This is beneficial, as a test sample with greater closeness to source domain ID samples intuitively suggests it is more likely to be ID itself, despite the domain shift. Secondly, this score increase means that this sample has greater influence, or weighting, when its score is propagated to its neighbors in the next iteration, thereby increasing the scores of semantically similar samples in the latent space. After several iterations, score information from reliable ID test samples will flow to their more uncertain ID neighbors, and vice versa for OOD samples, resulting in  more robust OOD scores.

\section{Experiments}
We conduct experiments to study the effectiveness of our methodology in point cloud OOD detection under domain shift. In practical settings, we would assign all test samples with a score below a user-defined threshold as OOD. However, the aim of this work is to improve the OOD scoring process, therefore we use evaluation metrics that do not require the setting of a specific threshold. Namely, AUC (higher is better) and FPR95 (lower is better), which is the false positive rate at $95\%$ recall. We also conduct an ablation study to analyze the behavior of our methodology under different experimental settings.

\subsection{Datasets}

\begin{table*}[h!]
    \centering
    \setlength{\tabcolsep}{4.2pt}
    \renewcommand{\arraystretch}{1}
    \caption{AUC (higher is better) and FPR95 (lower is better) scores. Best/second best scores are highlighted in bold/underlined. Baselines under `\textit{Customized Model}' are taken from \cite{alliegro20223dos} and all train a model using ID source domain data. Baselines under `\textit{Pre-trained Model}' use the pre-trained features from ULIP-2.}
    \begin{tabular}{lcccccccc}
    \toprule
    & \multicolumn{2}{c}{SR1
    } & \multicolumn{2}{c}{SR2} & \multicolumn{2}{c}{Average} \\
    & AUC $\uparrow$ & FPR95 $\downarrow$ & AUC $\uparrow$ & \multicolumn{1}{c}{FPR95 $\downarrow$} & \multicolumn{1}{c}{AUC $\uparrow$} & FPR95 $\downarrow$ \\
        \midrule
        \textbf{\textit{Customized Model}} \\
        \midrule
        MSP \cite{hendrycks2016baseline} & 81.0 & 79.6 & 70.3 & 86.7 & 75.6 & 83.2 \\ 
        MLS \cite{vaze2021open} & 82.1 & 76.6 & 67.6 & 86.8 & 74.8 & 81.7 \\ 
        ODIN \cite{liang2017enhancing} & 81.7 & 77.3 & 70.2 & 84.4 & 76.0 & 80.8 \\ 
        Energy \cite{liu2020energy} & 81.9 & 77.5 & 67.7 & 87.3 & 74.8 & 82.4 \\ 
        GradNorm \cite{huang2021importance }&  77.6 & 80.1 & 68.4 & 86.3 & 73.0 & 83.2 \\ 
        ReAct \cite{sun2021react} & 81.7 & 75.6 & 67.6 & 87.2 & 74.6 & 81.4 \\ 
        NF \cite{alliegro20223dos} & 78.0 & 84.4 & 74.7 & 84.2 & 76.4 & 84.3 \\ 
        OE+mixup \cite{hendrycks2018deep} &  71.2 & 89.7 & 60.3 & 93.5 & 65.7 & 91.6 \\ 
        ARPL+CS \cite{chen2021adversarial} &  82.8 & 74.9 & 68.0 & 89.3 & 75.4 & 82.1 \\ 
        Cosine proto \cite{fontanel2021detecting} &  79.9 & 74.5 & 76.5 & 77.8 & 78.2 & 76.1 \\ 
        CE (L2) \cite{alliegro20223dos} & 79.7 & 84.5 & 75.7 & 80.2 & 77.7 & 82.3 \\ 
        SubArcFace \cite{deng2020sub}&  78.7 & 84.3 & 75.1 & 83.4 & 76.9 & 83.8 \\ 
        \midrule
        \textbf{\textit{Pre-trained Model}} \\
        \midrule
        MSP* \cite{hendrycks2016baseline}& 83.0 & 84.2 & 74.6 & 81.4 & 78.8 & 82.8 \\
        MLS* \cite{vaze2021open} & 81.0	& 79.4	&  83.2	& 62.7	& 82.1	& 71.0 \\
        Cosine Proto \cite{fontanel2021detecting} & 80.7	& 70.2	& 73.6	& 83.3 & 77.1 & 76.8 \\
        Mahalanobis \cite{lee2018simple} & 73.8 & 89.7 & 65.3 & 83.3 & 69.5 & 86.5 \\
        OpenPatch \cite{rabino2023openpatch} & 85.8  & \underline{54.4} & 71.6 & 74.1 & 78.7 & 64.3 \\
        ZS-SODA* (\textit{Ours}) & \underline{85.9} & 67.1 & \underline{87.1} & \underline{50.4} & \underline{86.5} & \underline{58.7} \\
        SODA (\textit{Ours})& \textbf{93.3} & \textbf{33.3} & \textbf{87.7} & \textbf{47.4} & \textbf{90.5} & \textbf{40.4} \\
        \bottomrule
        \multicolumn{3}{l}{*source-free methods}
    \end{tabular}
    \label{tab:s2r}
\end{table*}

We follow the experimental framework of \cite{alliegro20223dos} in our main experiments, using the most popular benchmark point cloud datasets. In particular, we use real samples from ScanObjectNN \cite{uy-scanobjectnn-iccv19} as our test data and ModelNet40 \cite{wu20153d} as our reference data. We split the object classes into the three subsets shown in Table \ref{tab:class_splits}. All of the classes in SR1 and SR2 are also in ModelNet40, whereas SR3 classes are not. As such, we conduct one experiment with SR1 as ID and SR2$\cup$SR3 as OOD classes, and another experiment with SR2 as ID and SR1$\cup$SR3 as OOD classes. The number of reference samples in each class ranges from 109 to 889, and the details can be found in the supplementary material. Following \cite{alliegro20223dos}, we randomly sample 1024 points from reference point clouds and 2048 points from test point clouds.

We also experiment with ModelNet-C \cite{ren2022modelnetc}, which contains corrupted versions of ModelNet40 data, with multiple types of corruptions which are commonly observed in real data, such as global and local noise and point dropout. In these experiments, we use the clean ModelNet40 ID samples as reference samples and the corrupted ModelNet-C samples as test samples. We use the strongest level of corruption in our experiments, as this represents the greatest degree of domain shift. We conduct two types of experiments, one with each of SR1 and SR2 from Table \ref{tab:class_splits} as the ID/OOD samples respectively.


\subsection{Baselines}

\begin{table}
    \centering    
    \setlength{\tabcolsep}{6pt}
    \renewcommand{\arraystretch}{1}
    \caption{AUC (higher is better) and FPR95 (lower is better) performance of all pre-trained methods for ScanObjectNN test samples. ZS-SODA and SODA without propagation are simply the initial scores. Avg. change shows the average change after propagation over all methods.}
    \begin{tabular}{lcccccc}
    \toprule
        ~ & \multicolumn{2}{c}{S1} & \multicolumn{2}{c}{S2} & \multicolumn{2}{c}{Average} \\
        ~ & AUC $\uparrow$ & FPR95 $\downarrow$ & AUC $\uparrow$ & FPR95 $\downarrow$ & AUC $\uparrow$ & FPR95 $\downarrow$ \\ 
        \midrule
       \multicolumn{7}{c}{\textbf{Without Propagation}} \\ 
       \midrule
        MSP & 83.0 & 84.2 & 74.7 & 81.4 & 78.8 & 82.8 \\ 
        Source Similarity & 86.6 & 58.5 & 79.2 & 71.1 & 82.9 & 64.8 \\
        Cosine Proto & 80.7 & 70.2 & 73.6 & 83.3 & 77.1 & 76.8 \\ 
        ZS-SODA & 81.0	& 79.4 & 83.2 &	62.7 & 82.1 & 71.0 \\ 
        SODA & 88.6 & 63.0 & 84.8 & 58.0 & 86.7 & 60.5 \\ 
        \midrule
       \multicolumn{7}{c}{\textbf{With Propagation}} \\ 
       \midrule
        MSP & 87.7 & 61.9 & 78.7 & 71.3 & 83.2 & 66.6 \\ 
        Source Similarity & 90.1 & 43.2 & 84.5 & 55.2 & 87.3 & 49.2 \\
        Cosine Proto & 82.5	& 56.8 & 74.2 & 76.2 & 78.3 &	66.5 \\  
        ZS-SODA & 85.9 & 67.1 & 87.1 & 50.4 & 86.5 & 58.7 \\ 
        SODA & \textbf{93.3} & \textbf{33.3} & \textbf{87.7} & \textbf{47.4} & \textbf{90.5} & \textbf{40.4} \\ 
        \midrule
        Avg. change & 4.0 & -19.4 & 2.9 & -10.0 & 3.5 & -14.7 \\
        \bottomrule
    \end{tabular}
    \label{tab:effect_prop_mdnet}
\end{table}

Under `\textit{Customized Model}', we adopt the methods presented in \cite{alliegro20223dos} as baselines. All of these methods use models that have been trained on exclusively ID data from the source domain. We present their PointNet++ results due to its superior performance. Under, `\textit{Pre-trained Model}', we adopt methods that use pre-trained models. OpenPatch \cite{rabino2023openpatch} and Mahalanobis \cite{lee2018simple}, which measures the Mahanalobis distance to the reference samples, use a single modality PointBERT backbone. We also implement MSP, MLS and Cosine Proto, which measures the maximum similarity to ID reference sample prototypes, using the same features extracted from the 3D VLM ULIP-2 (PointBERT backbone) as ZS-SODA and SODA. For hyper-parameters, we use $T = 5$, $\alpha = 0.2$ and $\eta = 0.02$ by default without tuning. The effect of hyper-parameter settings is shown in the ablation study. We run our experiments in PyTorch on an Nvidia A5000 24G GPU.

\subsection{Results}

\subsubsection{ScanObjectNN}

Table \ref{tab:s2r} shows the average AUC and FPR95 scores over three random trials, for ID classes SR1 and SR2 separately, as well as their average. The standard deviations are shown in the supplementary material. We see that pre-trained methods mostly out-perform the customized model methods. This can be explained by the difference in model architecture as well as the superior representation learning that results from larger scale pre-training. We also see that our methodology significantly improves performance over all baselines. We see that ZS-SODA and SODA achieve an average of $3.9$ and $8.5$ percentage points improvement in AUC score over the next best-performing baseline, and a -13.4 and -31.4 point reduction in FPR95 respectively. SODA is also computationally efficient; we show in the supplementary material that similarity graph construction and score propagation contribute only a tiny portion of the overall runtime compared to the feature extraction phase. In the supplementary material, we also conduct experiments using both reference and test samples from ShapeNet \cite{chang2015shapenet} (i.e. all synthetic) and both reference and test samples from ScanObjectNN (i.e. all real), and find a similar improvement from our methodology.

We also measure the impact of score propagation using the other OOD methods as the initial scores, namely MSP, Cosine Proto and the source similarity $d_{src}$ itself. Table \ref{tab:effect_prop_mdnet} shows that score propagation greatly improves performance, with an average improvement of 3.5 points in AUC and -14.7 points in FPR95 across all methods, which demonstrates the positive effect of score propagation on OOD detection. We also see that SODA consistently outperforms the other methods, both with and without propagation, which demonstrates the complementary benefits of accounting for the similarity to the source domain alongside text similarity in detecting domain-shifted OOD inputs.

\begin{table}
    \centering
    \setlength{\tabcolsep}{7.2pt}
    \renewcommand{\arraystretch}{1.1}
    \caption{Performance of ZS-SODA and SODA (and change in performance from inital scores before propagation) on corrupted ModelNet-C test samples.}
    \begin{tabular}{lcccc}
    \toprule
      &  \multicolumn{2}{c}{ZS-SODA} & \multicolumn{2}{c}{SODA} \\ 
        Corruption & AUC $\uparrow$ & FPR95 $\downarrow$ & AUC $\uparrow$ & \multicolumn{1}{c}{FPR95 $\downarrow$}  \\ 
    \midrule
        Add Global & 89.7 (+2.7) & 41.4 (-11.5) & 97.1 (+1.5) & 20.9 (-9.0)  \\ 
        Add Local & 90.0 (+5.5) & 52.4 (-11.8) & 95.7 (+3.9) & 23.3 (-26.9)  \\ 
        Dropout Global & 76.9 (+3.9) & 75.3 (-7.9) & 88.4 (+5.7) & 62.3 (-12.7)  \\ 
        Dropout Local & 76.6 (+3.2) & 74.9 (-6.6) & 86.7 (+3.5) & 58.9 (-11.8)  \\ 
        Jitter & 48.6 (-0.6) & 97.8 (+1.6) & 60.5 (+1.5) & 88.0 (-2.7)  \\ 
        Rotate & 86.2 (+3.1) & 55.5 (-6.8) & 95.4 (+2.7) & 35.9 (-8.4)  \\ 
        Scale & 87.6 (+1.8) & 35.4 (-13.2) & 96.5 (+1.1) & 23.5 (-0.4)  \\ 
    \midrule
        \textbf{Average} & \textbf{79.4 (+2.8)} & \textbf{61.8 (-8.0)} & \textbf{88.6 (+2.8)} & \textbf{44.7 (-10.3)}  \\ 
    \bottomrule
    \end{tabular}
    \label{tab:modelnetc}
\end{table}

\subsubsection{ModelNet-C}

Table \ref{tab:modelnetc} shows the average performance with ModelNet-C test samples. For brevity, we show the average results over both experimental setups for ZS-SODA and SODA, followed by the change in these metrics from the initial scores before score propagation (in parentheses). We see that our methodology consistently improves performance in both metrics across corruption types, except for ZS-SODA which declines slightly for the jitter corruption. On average, we see a $2.8$ point improvement in AUC for both ZS-SODA and SODA, and a $-8.0/-10.3$ improvement in FPR95 scores for ZS-SODA/SODA. The full results are shown in the supplementary.

\subsection{Ablation Study}

\begin{figure*}
    \centering
    \begin{minipage}{0.32\textwidth}
        \centering
        \includegraphics[width=\textwidth]{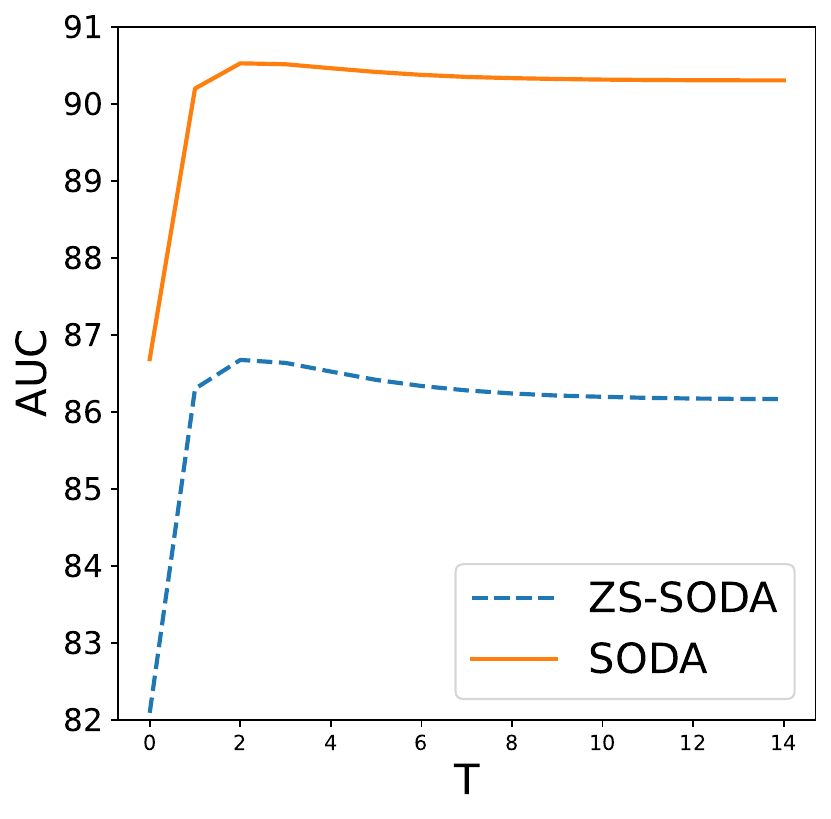}
    \end{minipage}
    \hfill
    \begin{minipage}{0.32\textwidth}
        \centering
        \includegraphics[width=\textwidth]{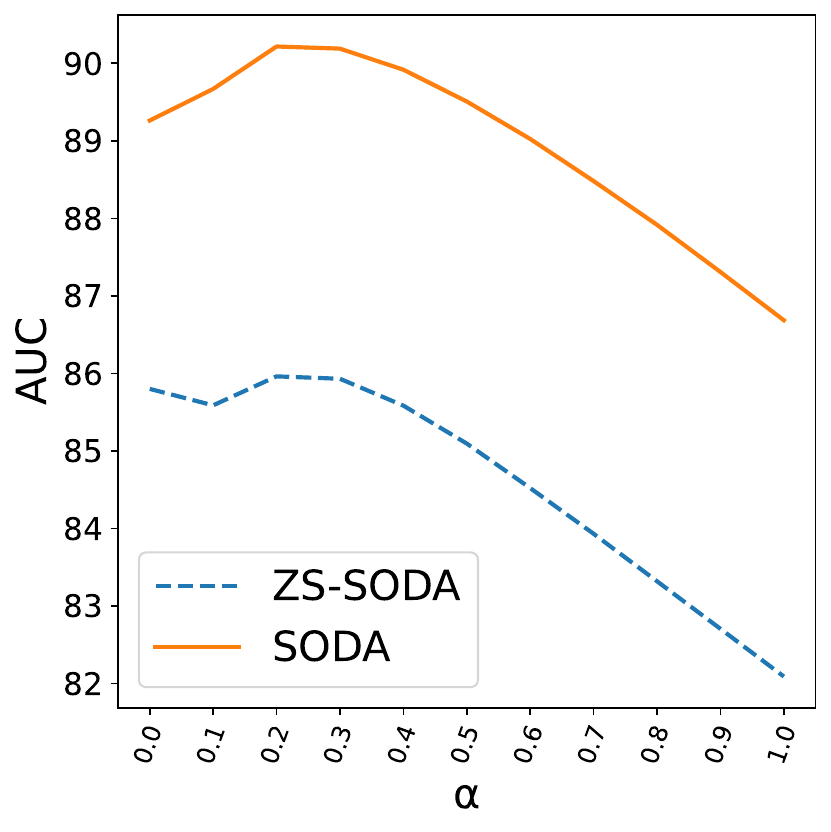}
    \end{minipage}
    \begin{minipage}{0.32\textwidth}
    \centering
    \includegraphics[width=\textwidth]{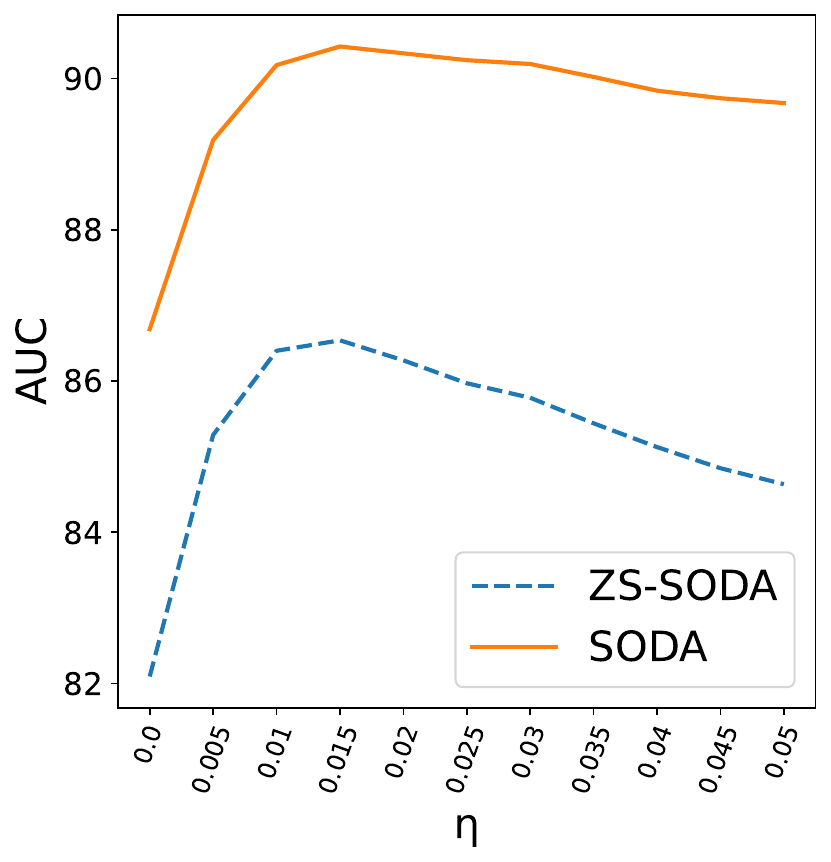}
    \end{minipage}
\caption{Average AUC performance with different numbers of iterations, $T$ (left), settings of $\alpha$ (middle), and settings of $\eta$ (right).}
\label{fig:ablation}
\end{figure*}

\subsubsection{Hyper-parameters}

Figure \ref{fig:ablation} shows how performance varies with different settings of hyper-parameters. Firstly, the leftmost figure shows AUC performance after different numbers of propagation iterations. $T=0$, gives the initial score, from which we see a large increase in performance after just one iteration. Performance slightly drops before converging at around $T = 6$ iterations, after which the OOD scores are fully converged and there is no more change in performance. In the middle figure, we see how performance changes as $\alpha$ in Eq. \ref{eq:propagation} varies. $\alpha = 1$ is equivalent to the initial scores, while a smaller $\alpha$ gives greater weight to the current score information from neighbors in score updates. We see that optimal performance is achieved around $\alpha \approx 0.25$, which signifies the importance of neighborhood information in refining the initial scores. The rightmost figure shows performance for different settings of $\eta$. A larger $\eta$ corresponds to a lower $\varepsilon$ threshold in Eq. \ref{eq:threshold}, and consequently a denser similarity graph with more connecting edges between samples.  We see peak performance is achieved at $\eta \approx 0.015$, and slowly declines from there. As $\eta$ increases, there are more samples propagating score information to neighbors of lower cosine similarity, which eventually hinders performance as the propagated information becomes less relevant. In practice, the setting of these hyper-parameters are guided by the number of available samples and also the sensitivity of individual applications to the trade-off between false positive and negative errors.

\subsubsection{Backbone models}

\begin{table}
    \centering
    \setlength{\tabcolsep}{4.3pt}
    \caption{AUC and FPR95 performance of SODA without propagation (the initial score) and with propagation for different backbone feature extraction models.}
    \begin{tabular}{lcccccc}
    \toprule
     & \multicolumn{2}{c}{Without Propagation} & \multicolumn{2}{c}{With Propagation} & \multicolumn{2}{c}{Avg. change} \\
    \midrule
    & AUC $\uparrow$ & FPR95 $\downarrow$ & AUC $\uparrow$ & FPR95 $\downarrow$ & AUC $\uparrow$ & FPR95 $\downarrow$ \\
     \midrule
     PointCLIPv2 & 63.7 & 91.4 & 65.3 & 86.5 & 1.6 & -4.9\\
     ULIP & 78.5 & 76.7 & 82.4 & 64.4 & 3.9 & -12.3\\
     ULIP-2 & \textbf{86.7} & \textbf{60.5} & \textbf{90.5} & \textbf{40.4} & \textbf{3.8} & \textbf{-20.1} \\
     \bottomrule
    \end{tabular}
    \label{tab:model}
\end{table}

Table \ref{tab:model} shows the performance of SODA before and after score propagation using different backbone 3D VLMs: ULIP-2, ULIP \cite{xue2023ulip} and PointCLIPv2 \cite{zhu2023pointclip}. We see that score propagation consistently improves both AUC and FPR95 metrics across all backbone models. Furthermore, we see that this improvement is most pronounced with ULIP-2, which also achieves the best performance overall. This aligns with its superior performance in other downstream tasks demonstrated in the original work. We can expect a better model to learn more meaningful latent representations, resulting in more semantically meaningful neighbors which are beneficial for score propagation.

In the supplementary material, we perform several additional experiments. Firstly, we test different formulations of text prompt templates and find that different prompts give better performance in different experimental setups. Overall, using the average of the embeddings from multiple prompt templates is more robust than any single template. 

\section{Conclusion}
We investigate the effect of synthetic-to-real domain shift on the embeddings learnt by pre-trained 3D vision-language models and find that the alignment of real domain point clouds with their corresponding text labels is degraded compared with synthetic data. This has significant implications for OOD detection and other practical downstream tasks that concern real point cloud data. To address this, we propose a novel methodology called SODA which updates and refines the OOD scores assigned to test samples using score information propagated from similar, nearby points within their local neighborhood. This has a local smoothening effect on the scoring function which improves robustness and significantly enhances detection performance. We achieve state-of-the-art performance in both AUC and FPR95 metrics across different experimental settings.

\begin{credits}
\subsubsection{\ackname} This research work is supported by the Agency for Science, Technology and Research (A*STAR) under its MTC Programmatic Funds (Grant No. M23L7b0021).

\end{credits}

\end{document}


\title{Supplementary Material}

\maketitle

\section{Dataset Statistics}

\begin{table}
    \setlength{\tabcolsep}{6pt}
    \renewcommand{\arraystretch}{1} 
    \caption{The number of reference samples for each ID class in ModelNet40.}
    \begin{tabular}{ccccccccccc}
    \toprule
     & bed & bookshelf & chair & desk & door & monitor & sink & sofa & table & toilet\\
        \midrule
     & 515 & 572& 889 & 200& 109& 465& 128& 680& 392& 344 \\
        \bottomrule
    \end{tabular}
    \label{tab:mdnet40_samples}
\end{table}

\section{Experiments}

\begin{table}[]
    \centering
    \caption{Runtime of each step of our methodology in seconds.}
    \setlength{\tabcolsep}{15pt}
    \renewcommand{\arraystretch}{1} 
    \begin{tabular}{ll}
    \toprule
    Step & Runtime (s) \\
    \midrule
    Feature extraction & 34.340582 \\
    Similarity graph construction & 0.112247 \\
    Initial Scoring & 0.235987\\
    Score propagation & 0.0012866$\backslash$iter \\
    \bottomrule
    \end{tabular}
    \label{tab:my_label}
\end{table}

\begin{table}
    \centering
     \setlength{\tabcolsep}{15pt}
    \renewcommand{\arraystretch}{1} 
    \caption{Standard deviation in AUC and FPR95 metrics of pre-trained methods over 3 random trials for the main experiments with ScanObjectNN.}
    \begin{tabular}{lcccc}
        \toprule
     & \multicolumn{2}{c}{SR1} & \multicolumn{2}{c}{SR2}\\
     &AUC & FPR95 & AUC & FPR95 \\
     \midrule
    MSP & 0.26&0.98&0.22&1.78 \\
    MLS & 0.24&0.642&0.27&1.94\\
    Cosine Proto &0.29&1.10&0.288&1.08\\
    ZS-SODA (\textit{Ours})&0.18&1.11&0.11&1.04\\
    SODA (\textit{Ours})&0.13&0.64&0.22&1.61\\
    \bottomrule
    \end{tabular}
    \label{tab:stdev}
\end{table}

\begin{table}[]
    \centering
    \caption{Full results for ModelNet-C experiments.}
    \begin{tabular}{lcccccc}
    \toprule
        ~ & \multicolumn{2}{c}{S1} & \multicolumn{2}{c}{S2} & \multicolumn{2}{c}{Average} \\ 
        ~ & AUC $\uparrow$ & FPR95 $\downarrow$ & AUC $\uparrow$ & FPR95 $\downarrow$ & AUC $\uparrow$ & FPR95 $\downarrow$ \\ 
        \midrule
        \multicolumn{7}{c}{ZS-SODA} \\
        Without SODA & ~ & ~ & ~ & ~ & ~ &   \\ 
        \midrule
        Add Global & 80.3 & 65.4 & 93.7 & 40.3 & 87.0 & 52.9  \\ 
        Add Local & 77.9 & 80.9 & 91.2 & 47.7 & 84.6 & 64.3  \\ 
        Dropout Global & 63.6 & 95.5 & 82.5 & 70.9 & 73.1 & 83.2  \\ 
        Dropout Local & 65.8 & 93.8 & 80.9 & 69.1 & 73.4 & 81.5  \\ 
        Jitter & 51.7 & 95.9 & 46.8 & 96.5 & 49.2 & 96.2  \\ 
        Rotate & 74.5 & 85.0 & 91.8 & 39.7 & 83.1 & 62.3  \\ 
        Scale & 77.5 & 68.7 & 94.0 & 28.5 & 85.7 & 48.6   \\ 
        Average & 70.2 & 83.6 & 83.0 & 56.1 & 76.6 & 69.8 \\ 
        \midrule
        With SODA & ~ & ~ & ~ & ~ & ~ &   \\ 
        \midrule
        Add Global & 83.6 & 40.1 & 95.7 & 42.7 & 89.7 & 41.4  \\ 
        Add Local & 83.9 & 75.7 & 96.1 & 29.1 & 90.0 & 52.4  \\ 
        Dropout Global & 64.8 & 98.2 & 89.1 & 52.4 & 76.9 & 75.3  \\ 
        Dropout Local & 66.5 & 91.8 & 86.6 & 57.9 & 76.6 & 74.9  \\ 
        Jitter & 51.0 & 98.8 & 46.3 & 96.8 & 48.6 & 97.8  \\ 
        Rotate & 77.8 & 76.3 & 94.5 & 34.7 & 86.2 & 55.5  \\ 
        Scale & 77.5 & 68.7 & 94 & 28.5 & 85.7 & 48.6  \\
        \midrule
        \midrule
        \multicolumn{7}{c}{SODA} \\
        Without Propagation & ~ & ~ & ~ & ~ & ~ & ~ \\ 
        \midrule
        Add Global & 95.5 & 26.8 & 95.8 & 32.9 & 95.6 & 29.8  \\ 
        Add Local & 90.1 & 56.4 & 93.3 & 44.1 & 91.7 & 50.3  \\ 
        Dropout Global & 79.8 & 81.5 & 85.7 & 68.5 & 82.7 & 75.0  \\ 
        Dropout Local & 80.1 & 84.8 & 86.4 & 56.8 & 83.3 & 70.8  \\ 
        Jitter & 60.8 & 97.7 & 57.2 & 83.5 & 59.0 & 90.6  \\ 
        Rotate & 90.4 & 61.7 & 95.2 & 26.8 & 92.8 & 44.2  \\ 
        Scale & 93.4 & 35.2 & 97.4 & 12.7 & 95.4 & 23.9  \\ 
        Average & 84.3 & 63.4 & 87.3 & 46.5 & 85.8 & 55.0  \\ 
        \midrule
        With Propagation & ~ & ~ & ~ & ~ & ~ &   \\ 
                \midrule
        Add Global & 97.1 & 19.1 & 97.1 & 22.7 & 97.1 & 20.9  \\ 
        Add Local & 94.5 & 29.0 & 96.9 & 17.7 & 95.7 & 23.3  \\ 
        Dropout Global & 85.9 & 73.7 & 91.0 & 50.9 & 88.4 & 62.3  \\ 
        Dropout Local & 82.4 & 76.1 & 91.1 & 41.8 & 86.7 & 58.9  \\ 
        Jitter & 65.0 & 99.2 & 55.9 & 76.8 & 60.5 & 88.0  \\ 
        Rotate & 94.4 & 43.2 & 96.5 & 28.5 & 95.4 & 35.9  \\ 
        Scale & 94.8 & 37.9 & 98.1 & 9.1 & 96.5 & 23.5  \\ 
        Average & 87.7 & 54.0 & 89.5 & 35.3 & 88.6 & 44.7  \\ 
        \bottomrule
    \end{tabular}
\end{table}

\begin{table*}
    \centering
    \renewcommand{\arraystretch}{0.85}
    \caption{Performance of the text score with different formulations of templates used in creating text prompts.}
    \begin{tabular}{lcccc|ccccc}
    \toprule
     & \multicolumn{4}{c}{SR1} & \multicolumn{4}{c}{SR2} \\ 
     & \multicolumn{2}{c}{Without} & \multicolumn{2}{c}{With Propagation} & \multicolumn{2}{c}{Without} & \multicolumn{2}{c}{With Propagation}  \\ 
     & AUC & FPR & AUC & \multicolumn{1}{c}{FPR} & \multicolumn{1}{c}{AUC} & FPR & AUC & FPR \\ 
     \midrule
    \midrule
    1. a photo of a CLS in the scene. & 81.01 & 72.48 & 86.56 & 57.00 & 77.76 & 80.21 & 83.07 & 65.94  \\ 
    2. a good photo of the CLS. & \textbf{82.36} & \textbf{71.93} & \textbf{89.73} & \textbf{50.40} & 79.87 & 70.88 & 85.94 & 54.61  \\ 
    3. a photo of the nice CLS. & 81.56 & 73.82 & 88.00 & 50.70 & 78.82 & 66.89 & 80.97 & 61.47  \\ 
    4. a photo of the large CLS. & 81.29 & 70.28 & 87.92 & 57.37 & 80.61 & 66.32 & 86.67 & 49.48  \\ 
    5. a bright photo of the CLS. & 79.01 & 77.86 & 83.44 & 67.58 & 79.03 & 73.60 & 83.55 & 62.04  \\ 
    6. a pixelated photo of a CLS. & 79.31 & 75.11 & 86.07 & 57.31 & 82.10 & 67.46 & 86.53 & 52.95  \\ 
    7. the cartoon CLS. & 72.93 & 88.44 & 81.94 & 75.11 & 79.16 & 74.83 & 87.50 & 51.57  \\ 
    8. an embroidered CLS. & 70.83 & 84.04 & 74.69 & 77.00 & 67.80 & 81.02 & 64.89 & 74.41  \\ 
    9. a painting of the CLS. & 69.85 & 93.70 & 71.56 & 90.58 & 77.02 & 75.83 & 77.31 & 71.60 \\ 
    \midrule
    1:9 (\textit{Ours}) & 81.26 & 78.72 & 86.11 & 66.79 & \textbf{82.92} & \textbf{64.89} & \textbf{86.74} & \textbf{51.57} \\
    \bottomrule
    \end{tabular}
    \label{tab:text prompts}
\end{table*}

\begin{table*}
    \centering
    \caption{AUC and FPR95 scores of full-shot OOD detection methods versus our few-shot method on the \textit{Synthetic} benchmark from \cite{alliegro20223dos}. Best/second best scores are highlighted in bold/underlined. All reference and test samples are from ShapeNet. SODA achieves the best performance in all settings.}
    \begin{tabular}{lcccccc|cc}
    \toprule
      & \multicolumn{2}{c}{SN1(hard)} & \multicolumn{2}{c}{SN2(med)} & \multicolumn{2}{c}{SN3(easy)} & \multicolumn{2}{c}{Average}\\
     & AUC $\uparrow$ & FPR95 $\downarrow$ & AUC $\uparrow$& FPR95 $\downarrow$ & AUC $\uparrow$ & \multicolumn{1}{c}{FPR95 $\downarrow$}  & AUC $\uparrow$ & \multicolumn{1}{c}{FPR95 $\downarrow$}\\
    \midrule
        MSP & 74.3 & 82.8 & 80.0 & 78.1 & 89.7 & 52.2 & 81.3 & 71.0 \\ 
        MLS & 72.0 & 80.8 & 83.9 & 64.1 & 89.8 & 40.5 & 81.9 & 61.8 \\ 
        ODIN & 74.2 & 79.4 & 79.4 & 71.7 & 87.8 & 41.8 & 80.5 & 64.3 \\ 
        Energy & 72.1 & 81.2 & 84.0 & 64.7 & 89.8 & 39.4 & 82.0 & 61.8 \\ 
        GradNorm & 72.1 & 81.8 & 57.7 & 88.9 & 57.8 & 79.0 & 62.6 & 83.3 \\ 
        ReAct & 73.7 & 79.4 & 89.6 & 52.1 & \underline{95.0} & \underline{27.2}& 86.1 & 52.9 \\ 
        $\mathrm{NF}$ & 81.5 & 72.5 & 71.1 & 78.0 & 91.0 & 49.6 & 81.2 & 66.7 \\ 
        OE+mixup & 72.7 & 78.9 & 80.3 & 68.8 & 87.3 & 62.2 & 80.1 & 69.9 \\ 
        ARPL+CS & 74.8 & 80.3 & 80.7 & 72.4 & 85.4 & 50.8 & 80.3 & 67.8 \\ 
        Cosine proto & 80.3 & 68.3 & 88.7 & 60.8 & 91.9 & 38.0 & 86.9 & 55.7 \\ 
        $\mathrm{CE}$ & 83.4 & 66.8 & 89.5 & 37.7 & 92.9 & 28.1 & 88.6 & 44.2 \\ 
        SubArcface & 79.0 & 81.2 & 82.9 & 60.3 & 89.1 & 32.8 & 83.7 & 58.1 \\ 
        \midrule
        MSP & 77.3	& 81.4	& 89.7& 52.1 & 87.2 & 57.9 & 84.7 & 63.8\\
        MLS &83.0&54.7&88.3&36.7&76.4&67.1&82.5&52.8\\
        Cosine Proto & \underline{82.0}&\underline{63.9}&82.3&59.6&91.3&38.9&\underline{85.2}&\underline{54.1}\\
        ZS-SODA (\textit{Ours}) & 85.0&42.4&\underline{90.9}&\underline{24.6}&85.0&55.7&87.0&40.9\\
        SODA (\textit{Ours})& \textbf{93.3}&\textbf{36.7}&\textbf{96.2}&\textbf{12.9}&\textbf{99.0}&\textbf{2.7}&\textbf{96.2}&\textbf{17.5} \\
        \bottomrule
        \end{tabular}
        \end{table*}
    
\begin{table*}[t!]
    \centering
    \caption{AUC and FPR95 scores on the \textit{Real} benchmark from \cite{alliegro20223dos}. Best/second best scores are highlighted in bold/underlined. Both reference and test samples are from ScanObjectNN. The customized models are trained on real samples, which explains the superior performance over the pre-trained ULIP-2 model methods, which was still pre-trained on synthetic data.}
    \begin{tabular}{lcccccc|cc}
    \toprule
        \midrule
        \multicolumn{9}{c}{\textit{Real} benchmark} \\
        \midrule
        ~ & \multicolumn{2}{c}{SR3 (easy)} & \multicolumn{2}{c}{SR2 (med)} &\multicolumn{2}{c}{SR1 (hard)} & \multicolumn{2}{c}{Average} \\ 
         & AUC $\uparrow$ & FPR95 $\downarrow$ & AUC $\uparrow$& FPR95 $\downarrow$& AUC $\uparrow$ & \multicolumn{1}{c}{FPR95 $\downarrow$} & \multicolumn{1}{c}{AUC $\uparrow$} & FPR95 $\downarrow$ \\
        \midrule
        MSP & 88.1 & 67.3 & 80.6 & 84.0 & 73.7 & 80.3 & 80.8 & 77.2 \\ 
        MLS & 89.4 & 53.8 & \textbf{83.4} & 73.1 & 76.4 & \underline{75.3} & 83.0 & 67.4 \\ 
        ODIN & 90.2 & \underline{47.9} & \underline{83.3} & 71.7 & 76.3 & 76.8 & \underline{83.3} & \underline{65.5} \\ 
        Energy & 89.5 & 50.6 & 81.6 & 75.8 & 76.6 & 75.5 & 82.6 & 67.3 \\ 
        GradNorm & 88.5 & 50.7 & 77.4 & 75.3 & 75.2 & 76.8 & 80.4 & 67.6 \\ 
        ReAct & \underline{90.3} & 48.9 & 82.4 & 75.8 & 75.4 & 77.6 & 82.7 & 67.4 \\ 
        NF & 88.0 & 47.7 & 80.6 & \underline{68.2} & 75.6 & 81.4 & 81.4 & 65.8 \\ 
        OE+mixup & 72.6 & 83.5 & 72.0 & 88.5 & 62.5 & 87.8 & 69.0 & 86.6 \\ 
        Cosine proto & \textbf{91.0} & \textbf{41.0} & 82.1 & 78.2 & 77.6 & 75.6 & \textbf{83.6} & \textbf{64.9} \\ 
        CE (L2) & 85.1 & 64.4 & 78.9 & 83.9 & 73.2 & 79.1 & 79.1 & 75.8 \\ 
        SubArcface & 87.1 & 61.3 & 78.9 & 76.9 & 73.7 & 81.4 & 79.9 & 73.2 \\ 
        \midrule
        MSP & 52.0	& 94.8 & 64.6 & 95.1 & 74.3 & 88.9 & 63.6	& 92.9\\
        MLS &67.5 & 78.1 & 63.7	& 95.7 & 79.4 & 78.6 & 70.2 & 84.1 \\
        Cosine Proto & 80.5 & 66.5 & 78.5 & 67.1 & 75.6 &	81.0 & 78.2 & 71.5\\
        ZS-SODA (\textit{Ours}) & 68.6&72.5&66.0&94.8&\underline{80.8}&84.1&71.8&83.8 \\
        SODA (\textit{Ours})&77.7&67.8&77.4&80.8&\textbf{89.7}&\textbf{75.0}&81.6&74.5\\
        \bottomrule
        \end{tabular}
        \label{tab:r2r}
        \end{table*}